%% file: main.tex
\documentclass[journal]{IEEEtran}
\ifCLASSINFOpdf
   \usepackage[pdftex]{graphicx}
\else
\fi

\input{math_commands.tex}

\usepackage{hyperref}
\usepackage{url}
\usepackage{amsmath}
\usepackage{amssymb}
\begin{document}
%
\title{Reducing Redundancy in the Bottleneck Representation of the Autoencoders}
%
%
%

\author{\IEEEauthorblockN{Firas Laakom\IEEEauthorrefmark{1} \thanks{The paper is under consideration at Pattern Recognition Letters},
Jenni Raitoharju\IEEEauthorrefmark{2}, Alexandros Iosifidis\IEEEauthorrefmark{3} and Moncef Gabbouj\IEEEauthorrefmark{1}}\\
\IEEEauthorblockA{\IEEEauthorrefmark{1}Department of Computing Sciences, Tampere University, Finland\\
\IEEEauthorrefmark{2}Programme for Environmental Information, Finnish Environment Institute, Finland\\
\IEEEauthorrefmark{3}Department of Electrical and Computer Engineering, Aarhus University, Denmark\\
Emails: \IEEEauthorrefmark{1}firas.laakom@tuni.fi,
\IEEEauthorrefmark{2}jenni.raitoharju@syke.fi,
\IEEEauthorrefmark{3}ai@ece.au.dk,
\IEEEauthorrefmark{1}moncef.gabbouj@tuni.fi}}

\maketitle

\begin{abstract}
Autoencoders are a type of unsupervised neural networks, which can be used to solve various tasks, e.g., dimensionality reduction, image compression, and image denoising. An AE has two goals: (i) compress the original input to a low-dimensional space at the bottleneck of the network topology using an encoder, (ii) reconstruct the input from the representation at the bottleneck using a decoder. Both encoder and decoder are optimized jointly by minimizing a distortion-based loss which implicitly forces the model to keep only those variations of input data that are required to reconstruct the and to reduce redundancies. In this paper, we propose a scheme to explicitly penalize feature redundancies in the bottleneck representation. To this end, we propose an additional loss term, based on the pair-wise correlation of the neurons, which complements the standard reconstruction loss forcing the encoder to learn a more diverse and richer representation of the input. We tested our approach across different tasks: dimensionality reduction using three different dataset, image compression using the MNIST dataset, and image denoising using fashion MNIST. The experimental results show that the proposed loss leads consistently to superior performance compared to the standard AE loss.
\end{abstract}


%
\IEEEpeerreviewmaketitle

\section{Introduction}
With the progress of data gathering techniques, high-dimensional data are available for training machine learning approaches. However, the impracticality of working in high dimensional spaces due to the \textit{curse of dimensionality} and the understanding that the data in many problems reside on manifolds with much lower dimensions than those of the original space, has led to the development of various approaches which try to learn a mapping from the original space to a more meaningful lower-dimensional representation.  

Autoencoders (AEs) \cite{goodfellow2016deep} are a powerful data-driven unsupervised approach used to learn a compact representation of a given input distribution. An autoencoder focuses solely on
finding a low dimensional representation, from which the input data can be reconstructed with minimal distortion. Autoencoders have been applied successfully in many applications, such as transfer learning \cite{deng2013sparse,zhuang2015supervised,kandaswamy2014improving}, anomaly detection \cite{beggel2019robust,zhao2017spatio,aygun2017network,zhou2017anomaly}, dimensionality reduction \cite{petscharnig2017dimensionality,thomas2016dimensionality,wang2014generalized}, and compression \cite{theis2017lossy,han2018deep,golinski2020feedback,yingzhen2018disentangled}.

To accomplish these tasks, an autoencoder has two different parts: an encoder $g(\cdot)$, which maps an input  $\vx \in \mathcal{X}$ to a compact low-dimensional space $g(\vx)$, called the bottleneck representation, and a decoder $f(\cdot)$, which takes the output of the encoder as its input and uses it to reconstruct the original input  $f \circ g(\vx)$. Given a distortion metric $D$: $\mathcal{X} \times \mathcal{X} \to \mathbb{R}$, which measures the difference between the original input and the reconstructed input \cite{pmlrv27baldi12a,deng2013sparse}, autoencoderss are trained in an end-to-end manner using gradient descent \cite{goodfellow2016deep} to minimize the loss $L$ defined as the average distortion over the training data $\{\vx_i\}_{i=1}^{N}$:
\begin{equation} 
\min_{f,g} L\big(\{\vx_i\}_{i=1}^{N}\big)  \triangleq \min_{f,g}  \frac{1}{N} \sum_{i=1}^N D(\vx_i,f\circ g(\vx_i)).
\end{equation}
Several extensions and regularization techniques have been proposed to augment this loss \cite{deng2013sparse,golinski2020feedback,theis2017lossy,cheng2018deep,seybold2019dueling} to improve the performance of the model. The extension are able to improve the mapping of the inputs to compressed compact representations at the bottleneck of the autoencoder, so that the original inputs can be better reconstructed from these compact representations using the decoder. 

By controlling the size of the bottleneck, one can explicitly control the dimensionality of the representation and the compression rate \cite{hu2020learning,theis2017lossy}. However, a low size of the bottleneck increases the complexity of the task of the decoder risking a higher distortion rate.  This trade-off forces the model to keep only those variations in the input data that are required to reconstruct the input and to avoid redundancies and noise within the input \cite{pmlrv27baldi12a,cheng2018deep}.  This is achieved implicitly using back-propagation by minimizing the  reconstruction error, i.e., distortion $D$.

\begin{figure}[t]
\centering
\includegraphics[width=0.8\linewidth]{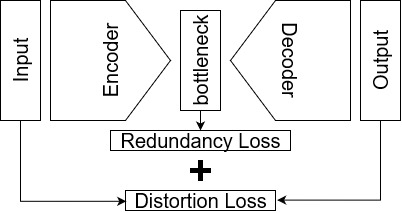}
\caption{An illustration on how the autoencoder loss is computed using our approach.}
\label{illustration}
\end{figure}

In the context of supervised neural networks, it has been shown that reducing the correlation improves generalization \cite{laakom2021within,cogswell2015reducing,laakom2021feature}. Approaches helping to reduce the redundancy have been successfully applied, e.g., for network pruning \cite{kondo2014dynamic, he2019filter, singh2020leveraging, lee2020filter} and self-supervised learning \cite{zbontar2021barlow}. In this paper, we propose to model the feature redundancy in the bottleneck representation and minimize it explicitly.  To this end, we propose augmenting the distortion loss L using a redundancy loss computed as the sum of the pair-wise correlations between the elements of the bottleneck. The full schema is illustrated in Figure \ref{illustration}. We argue that explicitly penalizing the cross-covariance between the different units in the bottleneck provides an extra feedback for the encoder to avoid redundancy and to learn a richer representation of the input samples.

The contributions of this paper can be summarized as follows: 
\begin{itemize}
    \item We propose a scheme to avoid redundant features in the bottleneck representation of autoencoders.
    \item We propose to augment the autoencoder loss to explicitly penalize the pair-wise correlations between the features and learn diverse compressed embedding of the samples.
    \item The proposed penalty acts as an unsupervised regularizer on top of the encoder and can be integrated into any autoencoder-based model in a plug-and-play manner.
    \item The proposed method is extensively evaluated over three tasks: dimensionality reduction, image compression, and image denoising. The results show a consistent performance boost compared to the standard approach.
\end{itemize}

The rest of this paper is organized as follows. Section \ref{related_work}
provides the background of autoencoders training strategies and a
brief review of different applications considered in this work: dimensionality reduction, image compression, and image denoising. Section \ref{proposed_approach} presents the proposed approach. Section \ref{dim_reduction} reports experimental results for dimensionality reduction task using three different datasets: Iris \cite{fisher1936use},  Cancer Wisconsin \cite{street1993nuclear}, and the Digit  \cite{alimoglu1996methods} Datasets.  Section \ref{im_compression} reports experimental results for the image compression task using the MNIST dataset \cite{lecun1998gradient}. Section \ref{image_denoising} evaluates our approach on the image denoising task using the fashionMNIST dataset \cite{xiao2017fashion}. Section \ref{conclusion} concludes the paper.

\section{Related work} \label{related_work}
Autoencoders were first introduced in \cite{rumelhart1985learning} as a model trained to reconstruct its input, i.e., to approximate the identity function $f(x) \approx x$.  While the identity function seems a particularly trivial function to learn,  enforcing certain constraints on the network topology and particularly using a low number of units in the hidden representations \cite{goodfellow2016deep} forces the model to learn to efficiently represent the data in a much lower-dimensional space compared to the original space \cite{baldi2012autoencoders}. This is a desired property in several applications, e.g., dimensionality reduction \cite{petscharnig2017dimensionality,thomas2016dimensionality,wang2014generalized}, compression \cite{theis2017lossy,han2018deep,golinski2020feedback,yingzhen2018disentangled}, and image denoising \cite{ye2015denoising,vincent2008extracting,gondara2016medical}. 

Dimensionality reduction refers to the problem of learning a mapping from a high-dimensional input space $\mathcal{X} \in \mathbb{R}^D$  into a lower-dimensional space $\mathcal{Z} \in \mathbb{R}^d$, where $ d <<D$, while preserving features of interest in the input data. Several linear \cite{zhao1999subspace,koren2004robust,laakom2020graph} and non-linear \cite{demers1993non,yeh2008nonlinear} approaches have been proposed to solve this task.  Some are supervised approaches \cite{liu2019supervised,sugiyama2006local}, such as Linear Discriminant Analysis (LDA) \cite{balakrishnama1998linear,li2020robust,iosifidis2013rvda} and Marginal Fisher Analysis (MFA) \cite{4016549}, others are unsupervised methods \cite{kumar2009analysis,dash1997dimensionality}, such as Principal Component Analysis (PCA) \cite{PCA1}. Dimensionality reduction is the most straightforward application of autoencoders \cite{petscharnig2017dimensionality,thomas2016dimensionality,wang2014generalized}. The desired mapping can be learned using an autoencoder by setting the size of the bottleneck to $d$ units and training the model to reconstruct a copy of the input.  


Image compression is an important task in many applications. Recent advances in deep neural networks \cite{goodfellow2016deep} have enabled efficient modeling of high-dimensional data and led to outperforming traditional compression techniques \cite{ullrich2017soft,mentzer2020learning,marcellin2000overview,skodras2001jpeg,rabbani1991digital} in image compression \cite{gregor2016towards,toderici2017full,balle2016end}. Recently, there has been interest in autoencoders to solve this problem \cite{ollivier2014auto,hu2020learning,cheng2018deep,theis2017lossy,rippel2017real} due to their flexibility and easiness to train \cite{theis2017lossy,hu2020learning,jiang2017end,yang2020improving}.

Image denoising \cite{gupta2013image} refers to the problem of trying to restore a clean version of the image from its noisy corrupted counterpart. Due to their plug-and-play network architectures, CNN-based autoencoders have been widely adopted to solve this task \cite{tian2020deep,goyal2020image}. In particular, an autoencoder is trained using pairs of noisy and clean images. By taking the noisy sample as an input and setting the clean version as a target at the output end, the model learns to keep only the important information from the image and discard the noise.

\section{Reducing the pair-wise correlation within the bottleneck representation} \label{proposed_approach}
Autoencoders are a special type of neural networks trained to achieve two objectives: (i) to compress an input signal into a low-dimensional space, (ii) to reconstruct the original input from the low-dimensional representation. This is achieved by minimizing the reconstruction loss over the training samples, which implicitly forces a concise `non-redundant' representation of the data. In this paper, we propose to augment the reconstruction  loss to explicitly minimize the redundancy, i.e., correlation, between the features learned at the bottleneck. 

Formally, given a training data $\{\vx_i\}_{i=1}^{N}$ and an encoder $g(\cdot) \in \mathbb{R}^D$, the correlation between the $i^{th}$ and $j^{th}$ features, $g_i$ and $g_j$, can be expressed as follows:
\begin{equation} 
C(g_i,g_j) = \frac{1}{N} \sum_n ( g_i(\vx_n)  - \mu_i ) ( g_j(\vx_n)  - \mu_j ),
\end{equation}
where $ \mu_i = \frac{1}{N} \sum_n g_i(\vx_n)  $ is the average output of the $i^{th}$ neuron. Our aim is to minimize the redundancy of the bottleneck representations which corresponds to minimizing the pair-wise covariance between different features. Thus, we augment the standard  loss  $L\big(\{\vx_i\}_{i=1}^{N}\big)$ as follows:
\begin{align} 
L\big(\{\vx_i\}_{i=1}^{N}\big)_{aug}  &\triangleq   L\big(\{\vx_i\}_{i=1}^{N}\big)  +\alpha  \sum_{i \neq j} C(g_i,g_j)\nonumber\\
&=  \frac{1}{N} \sum_{i=1}^N D(\vx_i,f\circ g(\vx_i)) \\
& +\alpha  \sum_{i \neq j} \Big( \frac{1}{N} \sum_n ( g_i(\vx_n)  - \mu_i ) ( g_j(\vx_n)  - \mu_j ) \Big),
\label{eqq}
\end{align}
where $\alpha$ is a hyper-parameter used to control the contribution of the additional term in the total loss of the model. $L_{aug}$ is composed of two terms, the first term is the traditional autoencoder loss that depends on both the encoder and decoder parts to ensure that the autoencoder learns to reconstruct the input, while the second term depends only on the encoder and its aim is to promote the diversity of the learned features and to ensure that the encoder learns less correlated non-redundant features.

Intuitively, the proposed approach acts as an unsupervised regularizer on top of the encoder providing an extra feedback during the back-propagation to reduce the correlations of the encoder's output. The proposed scheme can be embedded into any autoencoder-based model as a plug-in and optimized in a batch-manner, i.e., at each optimization step, we can compute the covariance using the batch samples. Moreover, it is suitable for different learning strategies and different topologies.

\section{Experiments on dimensionality reduction} \label{dim_reduction}
 
In this section, we consider the problem of dimensionality reduction using an autoencoder. We test the proposed approach using three different tabular dataset: Iris \cite{fisher1936use},  Cancer Wisconsin \cite{street1993nuclear}, and the Digits  \cite{alimoglu1996methods}. The important statistics of three datasets are summarized in Table \ref{datasets}. The Iris dataset \cite{fisher1936use} contains samples from three different Iris flower categories. Each sample is represented in a four-dimensional space and there is a total of $50 \times 3$ samples. The  Cancer Wisconsin dataset \cite{street1993nuclear}  is composed of 569 samples with features computed from a digitized image of a fine needle aspirate of a breast mass. These features describe characteristics of the cell nuclei present in the image. The Digits dataset \cite{alimoglu1996methods} is composed of  1797 $8\times8$ pixel images of digits. Each image sample is of a hand-written digit and its features are represented with a 64-dimensional vector.

\begin{table}[t]
\caption{Statistics of the three dataset}
\centering
\begin{tabular}{lccc}
\hline
Dataset & Dimensionality   & \# Classes   &      Samples per class       \\ 
\hline
IRIS \cite{fisher1936use} &  4 & 3 &  50 \\
Wisconsin \cite{street1993nuclear}  &  30 &  2 &  212-357\\
Digits \cite{alimoglu1996methods}  &  64 & 10 &   $\sim$180  \\
\hline
\end{tabular}
\label{datasets}	
\end{table}

As the autoencoder topology, we use a simple architecture, where the encoder maps the input using two intermediate fully-connected layers composed of 10 units with ReLU activation. Then, the bottleneck representation of size $d$ is obtained using a fully-connected layer with d units and Leaky ReLU \cite{maas2013rectifier} activation. Symmetrically, the decoder is composed of two 10-dimensional fully-connected layers followed by ReLU activation and an output layer with the same size as the input using a sigmoid activation. The value of $d$ is set to 2 for Iris and 6 for both Wisconsin and the Digit datasets. For training, we use Adam as our optimizer with a learning rate of $1\times 10^{-2}$ and the mean square error loss as our standard training loss $L$. The number of epochs and the batch size are set to 100 and 32 in  in all experiments, respectively. For all experiments, we use 75\% of the data for the training and the remaining 25\% for testing. Each experiment is repeated 5 times and the mean and standard deviation of the root mean square error (RMSE) on the test set are reported.  

In Table \ref{table_dr}, we report the experimental results using the standard loss and our proposed augmented loss using different values of hyper-parameter $\alpha$, introduced in \eqref{eqq}, in the range $\{0.1, 0.05,0.01,0.005,0.001\}$. It can be seen in the table that, by explicitly penalizing the redundancy in the bottleneck representation, our approach consistently achieves lower rates compared to the standard approach on the three datasets. 

\begin{table}[t]
\caption{Average and standard deviation of RMSE of different approaches on three different datasets over 5 repetitions}
\centering
\begin{tabular}{l|c|c|c}
\footnotesize
\label{table_dr}
& Iris & Wisconsin & Digits  \\
 \hline
standard & 0.0873 $\pm$ 0.0438  & 0.0717 $\pm$ 0.0056  & 0.1876 $\pm$ 0.0049 \\ 
ours (0.1)&  \textbf{0.0584} $\pm$ \textbf{0.0038} & 0.0704 $\pm$ 0.0047  & \textbf{0.1826} $\pm$ \textbf{0.0028} \\ 
ours (0.05) & 0.0609 $\pm$ 0.0090 & 0.0712 $\pm$  0.0048 & 0.1844 $\pm$ 0.0051 \\ 
ours (0.01) & 0.0656 $\pm$ 0.0154 & 0.0690 $\pm$  0.0062 & 0.1842 $\pm$ 0.0076 \\ 
ours (0.005) & \textbf{0.0584} $\pm$ \textbf{0.0028} & \textbf{0.0683} $\pm$ \textbf{0.0055}  & 0.1874 $\pm$ 0.0093  \\ 
ours (0.001) & 0.0795 $\pm$ 0.0466 & 0.0688 $\pm$  0.0053 &0.1859 $\pm$ 0.0037 \\ 
\hline
\end{tabular}
\end{table}

For the Iris dataset, the best performance is achieved using our approach with an $\alpha$ equal to 0.1 or 0.005, which improves the results on average $\sim33\%$ compared to the standard approach. For the Wisconsin dataset, using our approach with the hyperparameter $\alpha$ set to 0.005 achieves a 4.7\% average improvement compared to the standard approach, whereas, for the Digit dataset, the best performance is achieved using $\alpha=0.1$.

\section{Experiments on image compression} \label{im_compression}
In this section, we consider the problem of image compression using an autoencoder. We test the proposed approach using the MNIST dataset \cite{lecun1998gradient}, which is a handwritten digit dataset composed of 10 classes. MNIST images are $28 \times 28$ pixels, which results in 784-dimensional vectors. The dataset has 50000 samples for training and 10000 for testing.

For the autoencoder model, we use a simple architecture. The encoder is composed of two intermediate fully-connected layers composed of 256  and 128 units, respectively. The final output of the encoder is composed of $d$ units, where $d$ is the size of the bottleneck. Similarly, the decoder part takes the encoder's output, maps it to an intermediate layer of 128 units, then 256 units, and outputs a 784-vector. In all the layers, we use ReLU activation except for the final output, where sigmoid activation is used.

For the training, we use Adam as our optimizer with a learning rate of $1\times 10^{-2}$ and the mean square loss as our standard training loss $L$.  We train with 80\% of the original training data and hold the remaining 20\% as a validation set. During the training, the model with the lowest mean square error on the validation set is saved and used in the testing phase. We repeat each experiment five times and report the mean and standard deviation of the oot-mean-square error (RMSE) errors, the Peak signal-to-noise ratio
(PSNR), and structural index similarity (SSIM) scores on the test set for the different approaches. We experiment with two different bottleneck sizes, i.e., 256 and 64. The results for different bottleneck sizes are reported in Table \ref{table_compression}.

\begin{table}[t]
\caption{Average and standard deviation of RMSE, PSNR, and SSIM of different approaches on the MNIST dataset over 5 repetitions}
\centering
\begin{tabular}{l|c|c|c} \label{table_compression}
& RMSE  $ \downarrow$ & PSNR $ \uparrow$ & SSIM $ \uparrow$   \\
 \hline
& \multicolumn{3}{c}{784  $\to$  256 } \\
 \hline
Standard & 0.0518 $\pm$ 0.0005 & 0.9631 $\pm$ 0.0016  & 26.43 $\pm$ 0.09 \\ 
ours(0.1) & 0.0508 $\pm$ 0.0005 & 0.9641 $\pm$ 0.0010  & 26.57 $\pm$ 0.11 \\ 
ours(0.05) & 0.0508 $\pm$ 0.0005 & 0.9636 $\pm$ 0.0007  & 26.58 $\pm$ 0.08 \\ 
ours(0.01) & 0.0513 $\pm$ 0.0005 & \textbf{0.9647} $\pm$ \textbf{0.0013}  & 26.49 $\pm$ 0.09 \\ 
ours(0.005) & \textbf{0.0506} $\pm$ \textbf{0.0007} & 0.9635 $\pm$ 0.0012  & \textbf{26.61} $\pm$ \textbf{0.10} \\ 
\hline
& \multicolumn{3}{c}{784  $\to$  64 } \\
\hline
Standard & 0.0596 $\pm$ 0.0021 & 0.9597 $\pm$ 0.0022  & 25.25 $\pm$ 0.29 \\ 
Ours(0.1) & \textbf{0.0584} $\pm$ \textbf{0.0010} & \textbf{0.9607} $\pm$ \textbf{0.0012}  & \textbf{25.42} $\pm$ \textbf{0.16} \\ 
Ours(0.05) & 0.0588 $\pm$ 0.0018 & 0.9604 $\pm$ 0.0017  & 25.38 $\pm$ 0.25 \\ 
Ours(0.01) & 0.0593 $\pm$ 0.0010 & 0.9599 $\pm$ 0.0012  & 25.30 $\pm$ 0.15 \\ 
Ours(0.005) & 0.0588 $\pm$ 0.0009 & 0.9602 $\pm$ 0.0013  & 25.35 $\pm$ 0.13 \\ 
\end{tabular}
\end{table}

\begin{figure}[t]
\centering
\includegraphics[width=1\linewidth]{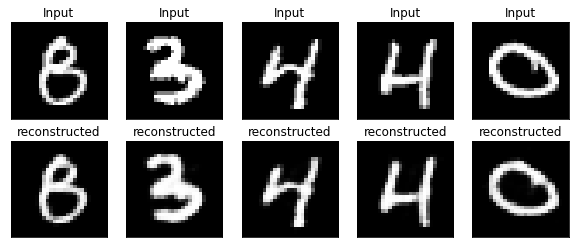}
\caption{An illustration of the performance of our approach on image compression. First row contains the original inputs and the second row contains their  reconstructed versions using our approach. }
\label{compression}
\end{figure}

We note that the proposed approach consistently boosts the performance of the autoencoder and yields lower RMSE errors compared to training with the standard loss only and higher PSNR and SSMI scores.  For $d=256$, the lowest RMSE score is achieved using the rate of 0.005 and the highest PSNR and SSMI scores are obtained by setting  $\alpha$ to 0.01 and 0.005, respectively. For $d=64$,   $\alpha$ equal to 0.1 corresponds to the best scores across all the metrics. Figure \ref{compression} presents visual outputs of images reconstructed from the feature representation learned by our approach for $d=64$. We note that our approach is able to learn to compress and  reconstruct the input without any major distortion to the input. 

\section{Experiments on image denoising} \label{image_denoising}

\begin{figure}[h]
\centering
\includegraphics[width=1\linewidth]{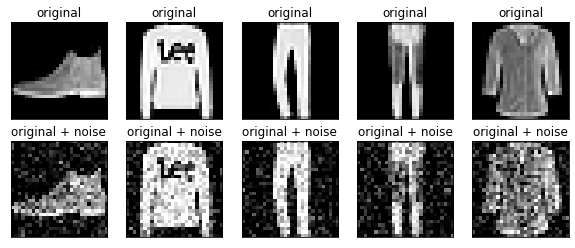}
\caption{An illustration of how the noise is added to 5 samples from fashionMNIST dataset with $\beta = 0.2$. First row contains the original samples and second row contains there corresponding noisy samples. }
\label{noisydata}
\end{figure}

In this section, we consider the problem of image denoising using an autoencoder.
We test the proposed approach using the fashionMNIST dataset \cite{xiao2017fashion}, which is an image dataset composed of 10 classes. Each sample is a $28 \times 28$ gray-scale image. The dataset has a total of  60,000 training samples and 10,000 test samples. To construct a noisy dataset, we add a random noise from normal distribution $\beta \times \mathcal{N}(0,1) $, where $\beta$ is a hyperparameter controlling the noise rate. In Figure \ref{noisydata}, we illustrate how the training samples are constructed from the original samples by adding the noise. 

As the autoencoder model, we use a simple CNN-based architecture. The encoder is composed of two convolutional layer $3\times3$ filters with sizes 16 and 4, respectively. Symmetrically, the decoder is composed of two transposed convolutional layers of sizes 4 and 16 and a final convolutional layer with one filter with kernel size $3\times3$. All the layers have ReLU activation function except for the last layer where we use  sigmoid activation. Each model is trained for 50 epochs using the mean square error loss and Adam optimizer. We repeat each experiment five times and report the means and standard deviations of RMSE, PSNR, and SSMI scores for different noise rates. 

\begin{table}[t]
\caption{Average and standard deviation of RMSE, PSNR, and SSMI scores of different approaches on the fashionMNIST dataset over 5 repetitions with different noise rates}
\centering
\begin{tabular}{l|c|c|c} \label{table_denosing}
& RMSE  $ \downarrow $  & PSNR  $ \uparrow$  & SSIM $ \uparrow$   \\
 \hline
& \multicolumn{3}{c}{$\beta=0.1$ } \\
 \hline
Standard & 0.0796 $\pm$ 0.0016 & 0.7980  $\pm$ 0.0061 & 22.51 $\pm$ 0.19  \\ 

Ours (0.1) & 0.0786 $\pm$ 0.0009 & 0.8018  $\pm$ 0.0024 & 22.64 $\pm$ 0.13  \\ 
Ours (0.05) & \textbf{0.0772} $\pm$ \textbf{0.0018} & 0.8049  $\pm$ 0.0062 & \textbf{22.84} $\pm$ \textbf{0.25}  \\ 
Ours (0.01) & 0.0779 $\pm$ 0.0019 & 0.8047  $\pm$ 0.0066 & 22.77 $\pm$ 0.27  \\ 
Ours (0.005) & 0.0774 $\pm$ 0.0012 & \textbf{0.8058}  $\pm$ \textbf{0.0044} & 22.82 $\pm$ 0.18  \\ 
\hline
& \multicolumn{3}{c}{$\beta=0.2$} \\
\hline
Standard & 0.0941 $\pm$ 0.0026 & 0.7283  $\pm$ 0.0110 & 20.95 $\pm$ 0.25  \\ 
Ours (0.1) & 0.0934 $\pm$ 0.0021 & 0.7301  $\pm$ 0.0102 & 21.03 $\pm$ 0.19  \\ 
Ours (0.05) & 0.0933 $\pm$ 0.0020 & 0.7290  $\pm$ 0.0079 & 21.04 $\pm$ 0.19  \\ 
Ours (0.01) & 0.0975 $\pm$ 0.0034 & 0.7143  $\pm$ 0.1276 & 20.63 $\pm$ 0.31  \\ 
Ours (0.005) & \textbf{0.0922} $\pm$\textbf{ 0.0012} & \textbf{0.7357}  $\pm$ \textbf{0.0058} & \textbf{21.14} $\pm$ \textbf{0.13}  \\ 
\hline
& \multicolumn{3}{c}{$\beta=0.4$} \\
\hline
Standard & 0.1262 $\pm$ 0.0021 & 0.5901  $\pm$ 0.0089 & 18.27 $\pm$ 0.16  \\ 
Ours (0.1) & \textbf{0.1258} $\pm$ \textbf{0.0021} & \textbf{0.5954}  $\pm$ \textbf{0.0095} & \textbf{18.30} $\pm$ \textbf{0.15}  \\ 
Ours (0.05) & 0.1260 $\pm$ 0.0016 & 0.5946  $\pm$ 0.0067 & 18.28 $\pm$ 0.12  \\ 
Ours (0.01) & 0.1266 $\pm$ 0.0014 & 0.5865  $\pm$ 0.0070 & 18.22 $\pm$ 0.09  \\ 
Ours (0.005) & 0.1260 $\pm$ 0.0017 & 0.5911  $\pm$ 0.0085 & 18.28 $\pm$ 0.13  \\ 
\hline
\end{tabular}
\end{table}

\begin{figure}[t]
\centering
\includegraphics[width=0.95\linewidth]{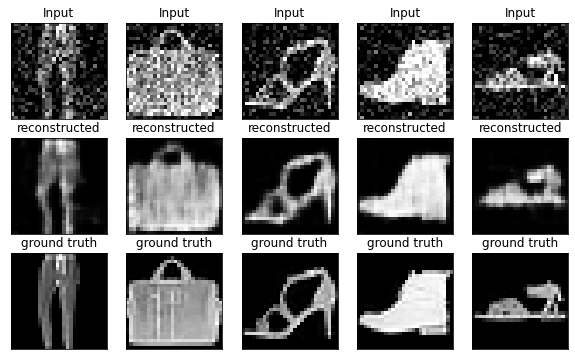}
\caption{Visual illustration of the performance of our approach on Image denoising. The first raw contains 5 input samples. Second raw and third raw present the output of our approach and the ground truth, respectively.}
\label{vizdenoiseoutput}
\end{figure}

In Table \ref{table_denosing}, we report the experimental results for three different noise rates $\beta$: 0.1, 0.2, and 0.4. Except for the hyper-parameter $\alpha=0.01$ with noise rates 0.2 and 0.4, we note that our approach by explicitly minimizing the redundancy in the feature space constantly outperforms the standard approach accross all metrics. For the noise rate $\beta=0.1$, the lowest RMSE rate and the highest SSMI score are achieved using our approach with $\alpha$ equal to 0.05, while the top PSNR is achieved with rates 0.005, whereas for a noise rate of 0.2, the best scores across all metrics correspond to $\alpha=0.005$. For the extreme noise case, i.e., $\beta = 0.4$, our approach with $\alpha=0.1$ achieves the best performance across the three metrics. In Figure \ref{vizdenoiseoutput}, we present visual outputs for our approach. As shown, our approach learns to efficiently discard the noise from the input images.

\section{Conclusion} \label{conclusion}
In this paper, we proposed a schema for modeling the redundancies at the bottleneck of an autoencoder. We proposed to complement the loss with an extra regularizer, which explicitly  penalizes the pair-wise correlation of the neurons at the encoder's output and, thus, forces it to learn more diverse and compact representations for the input samples. The proposed approach can be interpreted as an unsupervised regularizer on top of the encoder and can be integrated into any autoencoder-based model in a plug-and-play manner. We empirically demonstrated the effectiveness of our approach across multiple tasks, dimensionality reduction, compression, and denoising ,and we showed that it boosts the performance compared to the standard approach. Future directions include extensive testing of our approach with deep autoencoders and proposing more efficient redundancy modeling techniques to improve the performance of autoencoders.

\section*{Acknowledgment}
This  work  has been  supported  by  the  NSF-Business  Finland
Center for Visual and Decision Informatics (CVDI) project
AMALIA. The work of Jenni Raitoharju was funded by the Academy of Finland (project 324475). 

\ifCLASSOPTIONcaptionsoff
  \newpage
\fi



\bibliographystyle{IEEEtran}
\bibliography{main.bib}

\end{document}

%% file: math_commands.tex
\usepackage{amsmath,amsfonts,bm}

\def\eqref#1{equation~\ref{#1}}

\def\1{\bm{1}}

\def\vx{{\bm{x}}}

\DeclareMathAlphabet{\mathsfit}{\encodingdefault}{\sfdefault}{m}{sl}
\SetMathAlphabet{\mathsfit}{bold}{\encodingdefault}{\sfdefault}{bx}{n}

